\relax
\documentclass[letterpaper]{article}
\usepackage{aaai17_mod}
\usepackage{times}
\usepackage{helvet}
\usepackage{courier}
\usepackage{epsfig}
\usepackage{graphicx}
\usepackage{amsmath}
\usepackage{amssymb}

\usepackage{array}
\usepackage{color}

\newcommand\relphantom[1]{\mathrel{\phantom{#1}}}
\DeclareMathOperator*{\argmax}{arg\,max}
\DeclareMathOperator*{\argmin}{arg\,min}

\begin{document}

\title{Privacy-Preserving Human Activity Recognition from Extreme Low Resolution}

\author{Michael S. Ryoo$^{1,2}$, Brandon Rothrock$^3$, Charles Fleming$^4$, Hyun Jong Yang$^{2,5}$\\
$^1$Indiana University, Bloomington, IN\\
$^2$EgoVid Inc., Ulsan, South Korea\\
$^3$Jet Propulsion Laboratory, California Institute of Technology, Pasadena, CA\\
$^4$Xi'an Jiaotong-Liverpool University, Suzhou, China\\
$^5$Ulsan National Institute of Science and Technology, Ulsan, South Korea\\
\texttt{mryoo@indiana.edu}\\
}


\maketitle

\begin{abstract}
Privacy protection from surreptitious video recordings is an important societal challenge. We desire a computer vision system (e.g., a robot) that can recognize human activities and assist our daily life, yet ensure that it is not recording video that may invade our privacy. This paper presents a fundamental approach to address such contradicting objectives: human activity recognition while only using extreme low-resolution (e.g., 16x12) anonymized videos. We introduce the paradigm of \emph{inverse super resolution (ISR)}, the concept of learning the optimal set of image transformations to generate multiple low-resolution (LR) training videos from a single video. Our ISR learns different types of sub-pixel transformations optimized for the activity classification, allowing the classifier to best take advantage of existing high-resolution videos (e.g., YouTube videos) by creating multiple LR training videos tailored for the problem. We experimentally confirm that the paradigm of inverse super resolution is able to benefit activity recognition from extreme low-resolution videos.
\end{abstract}

\begin{figure}
	\centering
	\resizebox{0.86\linewidth}{!}{
	  \includegraphics{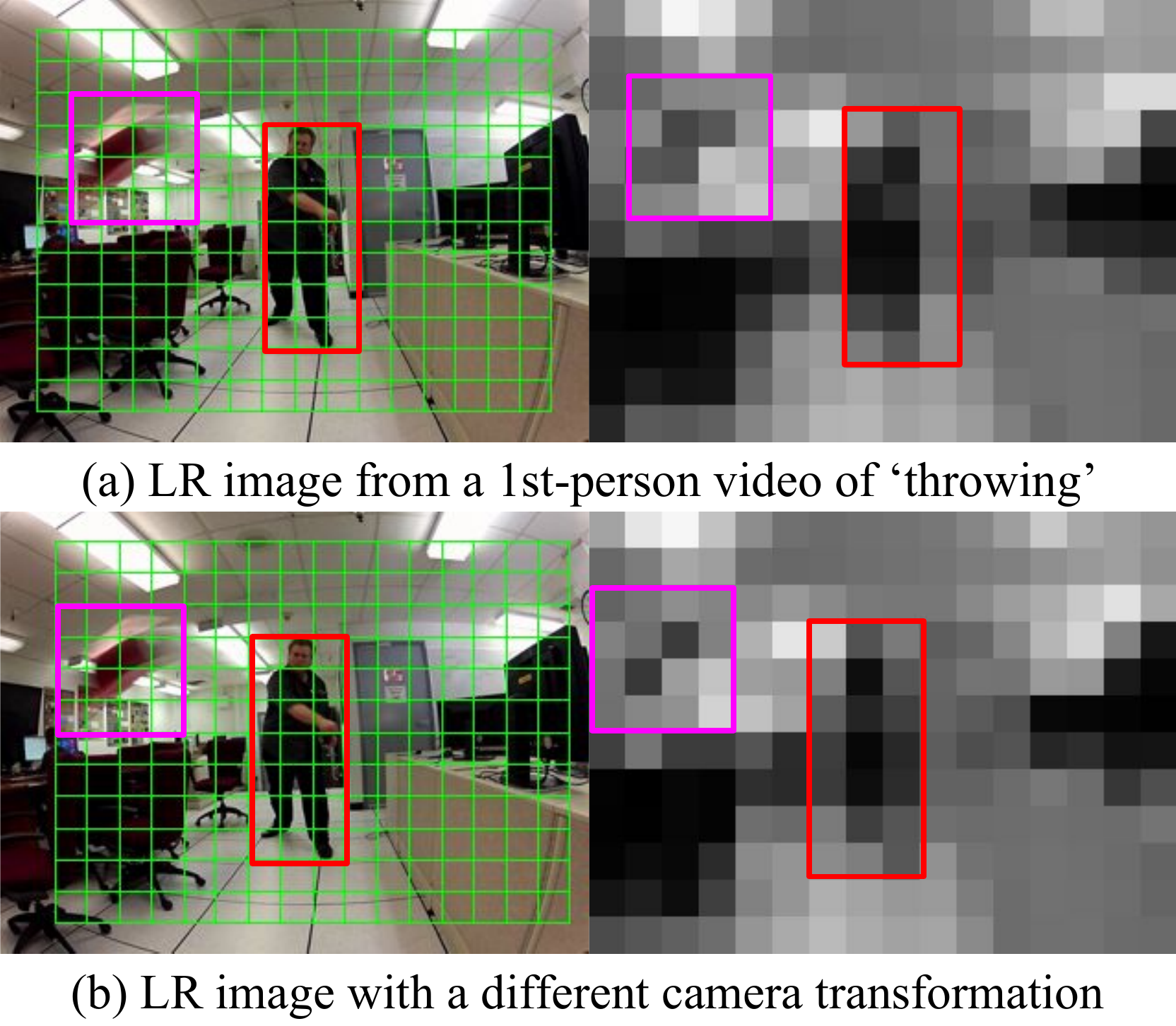}
	}
	\caption{Even though LR images in both (a) and (b) are from the same original scene, because of the inherent limitation of pixels, the two LR images show very different visual structures when the camera projection is slightly different. The pixel values corresponding to objects (red box: a human / magenta box: a can) differ significantly in these two LR images, making their visual features also different.}
	\label{fig:isr-motivation}
\end{figure}

\section{Introduction}

Cameras are becoming increasingly ubiquitous and pervasive. Millions of surveillance cameras are recording people's everyday behavior at public places, and people are using wearable cameras designed for lifelogging (e.g., GoPro and Narrative Clips) to obtain large collections of egocentric videos. Furthermore, robots at public places are equipped with multiple cameras for their operation and interaction.


Simultaneously, such abundance of cameras is also causing a big societal challenge: privacy protection from unwanted video recordings. We want a camera system (e.g., a robot) to recognize important events and assist human daily life by understanding its videos, but we also want to ensure that it is not intruding the user's or others' privacy. This leads to two contradicting objectives. More specifically, we want to (1) prevent the camera system from obtaining detailed visual data that may contain private information (e.g., faces), desirably at the hardware-level. Simultaneously, we want to (2) make the system capture as much detailed information as possible from its video, so that it understands surrounding objects and ongoing events for surveillance, lifelogging, and intelligent services.

There have been previous studies corresponding to such societal needs. Templeman et al. \shortcite{templeman14} studied scene recognition from images captured with wearable cameras, detecting locations where the privacy needs to be protected (e.g., restrooms). This will allow the device to be automatically turned off at sensitive locations. One may also argue that limiting the device to only process/transfer feature information (e.g., HOG and CNN) instead of visual data will make it protect privacy. However, recent studies on feature ``visualizations'' \cite{vondrick15} showed that it actually is possible to recover a good amount of visual information (i.e., images and videos) from the feature data. Furthermore, all these methods described above rely on software-level processing of original high-resolution videos (which may already contain privacy sensitive data), and there is a possibility of these original videos being snatched by cyber attacks.

A more fundamental solution toward the construction of a privacy-preserving vision system is the use of \emph{anonymized videos}. Typical examples of anonymized videos are videos made to have extreme low resolution (e.g., 16x12) by using low resolution (LR) camera hardware, or based on image operations like Gaussian blurring and superpixel clustering \cite{butler15}. Instead of obtaining high-resolution videos and trying to process them, this direction simply limits itself to only obtain anonymized videos. The idea is that, if we are able to \textbf{develop reliable computer vision approaches that only utilize such anonymized videos}, we will be able to do the recognition while preserving privacy. Such a concept may even allow cameras that can intelligently select their resolution; it will use high-resolution cameras only when it is necessary (e.g., emergency), determined based on extreme low-resolution video analysis.

There were previous attempts under such paradigm: \cite{dai15}. This conventional approach was to resize the original training videos to fit the target resolution, making training videos to visually look similar to the testing videos. However, there is an intrinsic problem: because of natural limitations of what a pixel can capture in a LR video, features being extracted from LR videos change a lot depending on sub-pixel camera viewpoint changes even when they contain the exact same object/human: Figure \ref{fig:isr-motivation}. This makes the decision boundary learning unstable.







In this paper, we introduce the novel concept of \emph{inverse super resolution} to overcome such limitations. Super resolution reconstructs a single high-resolution image from a set of low-resolution images. Inverse super resolution is the reverse of this process: we learn to generate a set of informative LR images from one HR image. Although it is natural to assume that the system obtains only one low-resolution `testing' video for privacy protection, in most cases, the system has an access to a rich set of high-resolution `training' videos publicly available (e.g., YouTube). The motivation behind inverse super resolution is that, if it really is true that a set of low-resolution images contains comparable amount of information to a high-resolution image, then we can also generate a set of LR training images from a HR image so that the amount of training information is maintained. Our approach learns the optimal set of LR transformations to make such generation possible, and uses the generated LR videos to obtain LR decision boundaries (Figure \ref{fig:isr}).





\section{Related works}

Human activity recognition is a computer vision area with a great amount of attention \cite{aggarwal11}. There not only have been studies to recognize activities from YouTube-style videos \cite{google15} or surveillance videos, but also first-person videos from wearable cameras \cite{kitani11,ramanan12,lee12,li13,poleg14} and robots \cite{ryoo13}. However, they only focused on developing features/methods for more reliable recognition, without any privacy aspect consideration.



\begin{figure}
	\centering
	\resizebox{1.0\linewidth}{!}{
	  \includegraphics{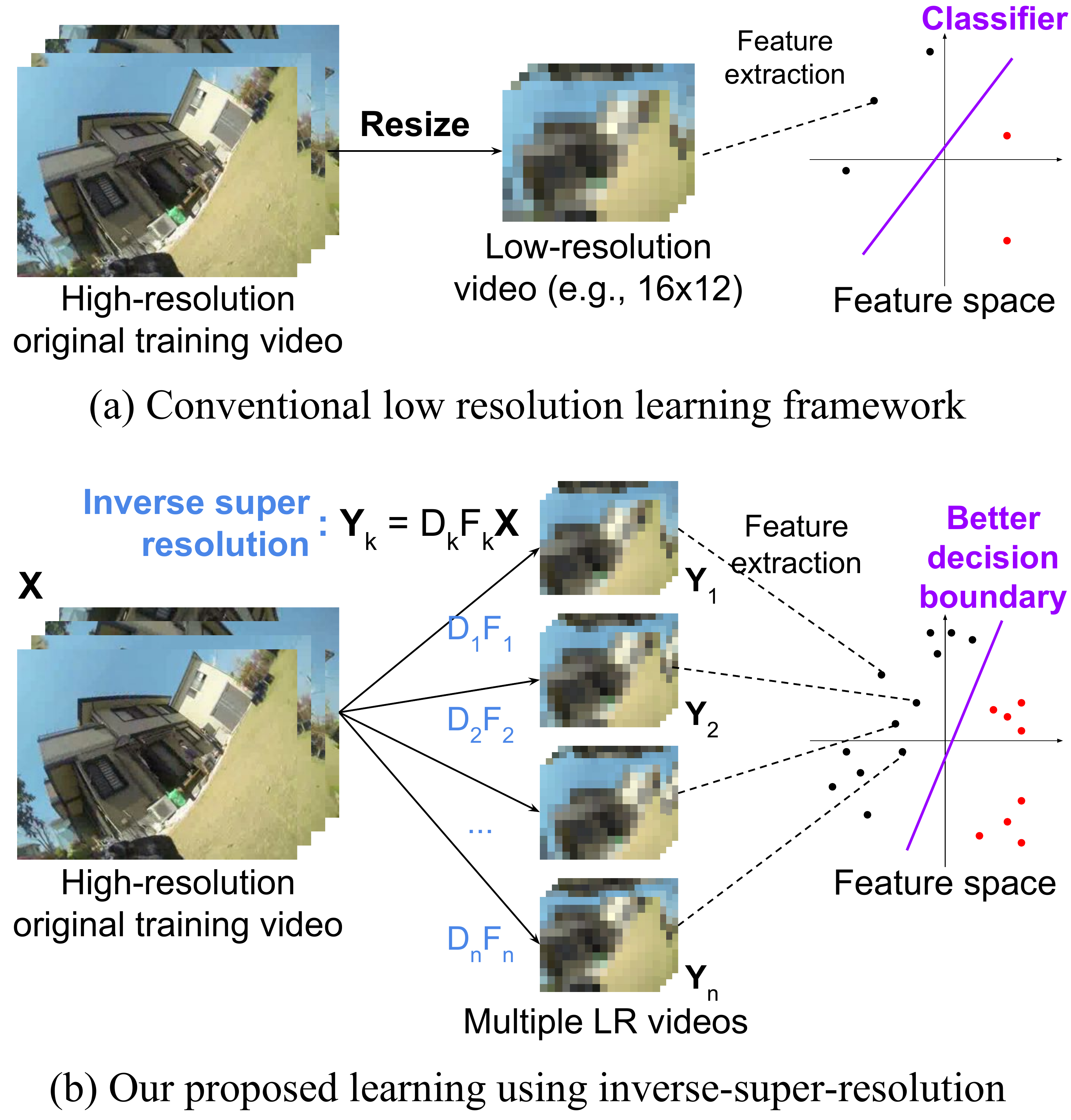}
	}
	\caption{A figure comparing (a) the conventional learning framework for low-resolution videos vs. (b) our learning framework using the proposed inverse super resolution.}
	\label{fig:isr}
\end{figure}

On the other hand, as mentioned in the introduction, there are research works whose goal is to specifically address privacy concerns regarding unwanted video taking. Templeman et al. \shortcite{templeman14} designed a method to automatically detect locations where the cameras should be turned off. \cite{tran16} was similar. Dai et al. \shortcite{dai15} studied human activity recognition from extreme low resolution videos, different from conventional activity recognition literature that were focusing mostly on methods for images and videos with sufficient resolutions. Although their work only focused on recognition from 3rd-person videos captured with static cameras, they showed the potential that computer vision features and classifiers can also work with very low resolution videos. However, they followed the `conventional paradigm' described in Figure \ref{fig:isr} (a), simply resizing original training videos to make them low resolution. \cite{shokri15} studied privacy protection for convolutional neural networks (CNNs), but they consider the privacy protection only at the training phase and not at the testing phase, unlike ours.



\section{Inverse super resolution}
\label{sec:isr}

\emph{Inverse super resolution} (ISR) is the concept of generating a set of low-resolution training images from a single high-resolution image, by `learning' different image transforms optimized for the recognition task. Such transforms may include sub-pixel translation, scaling, rotation, and other affine transforms emulating possible camera motion.

Our ISR targets the realistic scenario where the system is prohibited from obtaining high-resolution videos in the \emph{testing} phase due to privacy protection but has an access to a rich set of high-resolution training videos publicly available (e.g., YouTube). Instead of trying to enhance the resolution of the testing video (which is not possible with our scale factor x20), our approach is to make the system learn to benefit from high-resolution training videos by imposing multiple different sub-pixel transformations. This enables us to better estimate the decision boundary in the low-resolution feature space, as illustrated in Figure \ref{fig:isr}.

From the super resolution perspective, this is a different way of using the super resolution formulation, whose assumption is that multiple LR images may contain a comparable amount of information to a single HR image. It is called `inverse' super resolution since it follows the super resolution formulation, while the input and the output is the reverse of the original super resolution.



\subsection{Inverse super resolution formulation}
The goal of the super resolution process is to take a series of low resolution images $Y_k$, and generate a high resolution output image $X$ \cite{huang10}. This is done by considering the sequence of images $Y_k$ to be different views of the high resolution image $X$, subject to camera motion, lens blurring, down-sampling, and noise. Each of these effects is modeled as a linear operator, and the sequence of low resolution images can be written as a linear function of the original high resolution image: 
\begin{equation}
Y_k = D_k H_k F_k X + V_k, ~~k = 1 \ldots n
\end{equation}
where $F_k$ is the motion transformation, $H_k$ models the blurring effects, $D_k$ is the down-sampling operator, and $V_k$ is the noise term. $X$ and $Y_k$ are both images flattened into 1D vectors. In the original super resolution problem, none of these operators are known exactly, resulting in an ill-posed, ill-conditioned, and most likely rank deficient problem. Super resolution research focuses on estimating these operators and the value of $X$ by adding additional constraints to the problem, such as smoothness requirements.


Our inverse super resolution formulation can be described as its inverse problem. We want to generate multiple (i.e., $n$) low resolution images (i.e., $Y_k$) for each high resolution training image $X$, by applying the optimal set of transforms $F_k$, $H_k$, and $D_k$. We can simplify our formulation by removing the noise term $V_k$ and the lens blur term $H_k$, since there is little reason to further contaminate the resulting low-resolution images and confuse the classifiers in our case:
\begin{equation}
\label{eq:isr}
Y_k = D_k F_k X, ~~k = 1 \ldots n.
\end{equation}
In principle, $F_k$, the camera motion transformation, can be any affine transformation. We particularly consider combinations of shifting, scaling, and rotation transforms as our $F_k$. We use the standard average downsampling as our $D_k$.

The main technical challenge here is that we need to learn the set of different motion transforms $S = \{F_k\}^n_{k=1}$ from a very large pool, which is expected to maximize the recognition performance when applied to generate the training set for classifiers. Such learning should be done based on training data and should be dependent on the features and the classifiers being used, solving the optimization problem in the feature space.


Once $S$ is learned, the inverse super resolution allows generation of multiple low resolution images $Y_k$ from a single high resolution image $X$ by following Equation \ref{eq:isr}. Low resolution `videos' can be generated in a similar fashion to the case of images. We simply apply the same $D_k$ and $F_k$ for each frame of the video $X$, and concatenate the results to get the new video $Y_k$.

\subsection{Recognition with inverse super resolution}

Given a set of transforms $S = \{F_k\}^n_{k=1}$, the recognition framework with our inverse super resolution is as follows: For each of original high resolution training video $X_i$, we apply Equation \ref{eq:isr} to generate $n$ number of $Y_{ik} = D_k F_k X_i$. Let us denote the ground truth label (i.e., activity class) of the video $X_i$ as $y_i$. Also, we describe the feature vector of the video $Y_{ik}$ more simply as $x_{ik}$. The features extracted from all LR videos generated using inverse super resolution become training data. That is, the training set $T_n$ can be described as $T(S) = \cup_i\{ \langle x_{ik}, y_i \rangle \}_{k=0}^n$ where $n$ is the number of LR training samples to be generated per original video. $x_{i0}$ is the original sample resized to LR as is.

Based on $T(S)$, a classification function $f(x)$ with the parameters $\theta$ is learned. The proposed approach can cope with any types of classifiers in principle. In the case of SVMs with the non-linear kernels we use in our experiments,
\begin{equation}
	\begin{aligned}
		f_{\theta}(x) = \sum_j \alpha_j y_j K(x, x_j) + b
	\end{aligned}
\end{equation}
where $\alpha_j$ and $b$ are SVM parameters, $x_j$ are support vectors, and $K$ is the kernel function being used.


\section{Transformation learning}
\label{sec:learning}

In the previous section, we presented a new framework that takes advantage of LR training videos generated from HR videos assuming a `given' set of transforms. In this section, we present methods to `learn' the optimal set of such motion transforms $S = \{F_k\}^n_{k=1}$ based on video data. Such $S$ learned from video data is expected to perform superior to transforms randomly selected or uniformly selected, which we further confirm in our experiments.

\subsection{Method 1 - decision boundary matching}

Here, we present a Markov chain Monte Carlo (MCMC)-based search approach to find the optimal set of transforms providing the ideal activity classification decision boundaries. The main idea is that, if we have an infinite number of transforms $F_k$ generating LR training samples, we would be able to learn the best low-resolution classifiers for the problem. Let us denote such ideal decision boundary as $f_{\theta^*}$. By trying to minimize the distance between $f_{\theta^*}$ and the decision boundary that can be learned with our transforms, we want to find a set of transformations $S^*$:
\begin{equation}
    \label{eq:boundary-sim}
	\begin{aligned}
		S^* &= \argmin_{S} \left| f_{\theta^*} - f_{\theta(S)} \right|\\
		    & \approx \argmin_{S} \sum_{x \in A} \left| f_{\theta^*}(x) - f_{\theta(S)}(x) \right|\\
		\textrm{s.t.}~~ & |S^*| = n
	\end{aligned}
\end{equation}
where $f_{\theta(S)}(x)$ is a classification function (i.e., a decision boundary) learned from the training set $T(S)$ (i.e., LR videos generated using transforms in $S$). $A$ is a validation set with activity videos, being used to measure the empirical similarity between two classification functions.

In our implementation, we further approximate the above equation, since learning $f_{\theta^*}(x)$ conceptually requires an infinite (or very large) number of transform filters $F_k$. That is, we assume $f_{\theta^*}(x) \approx f_{\theta(S_L)}(x)$ where $S_{L}$ is a set with a large number of transforms. We also use $S_{L}$ as the `pool' of transforms we consider: $S \subset S_L$.


We take advantage of a MCMC sampling method with Metropolis-Hastings algorithm, where each MCMC action is adding or removing a particular motion transform filter $F_k$ to/from the current set $S^{t}$. The transition probability $a$ is defined as 
\begin{equation}
	\begin{aligned}
		a = \frac{\pi(S') \cdot q(S', S^{t})}{\pi(S^{t}) \cdot q(S^{t}, S')}
	\end{aligned}
\end{equation}
where the target distribution $\pi(S)$ is computed by
\begin{equation}
	\begin{aligned}
		\pi(S) \propto e^{-\sum_{x \in A} \left| f_{\theta^*}(x) - f_{\theta(S)}(x) \right|},
	\end{aligned}
\end{equation}
which is based on the $\argmin$ term of Equation \ref{eq:boundary-sim}. We model the proposal density $q(S', S^{(t)})$ with a Gaussian distribution $|S'| \sim N(n, \sigma^2)$ where $n$ is the number of inverse super resolution samples we are targeting (i.e., the number of filters). The proposal $S'$ is accepted with the transition probability $a$, and it becomes $S^{t+1}$ once accepted.

Using the above MCMC formulation, our approach goes through multiple iterations from $S^0 = \{ \}$ to $S^m$ where $m$ is the number of maximum iterations we consider. Based on the sampled $S^0, \ldots, S^m$, the one with the maximum $\pi(S)$ value is finally chosen as our transforms: $S^* = \argmax_{S^t} \pi(S^t)$ with the condition $|S| \leq n$.

\subsection{Method 2 - maximum entropy}

In this subsection, we propose an alternative methodology to learn the optimal set of transform filters $S^*$. Although the above methodology of directly comparing the classification functions provides us a good solution for the problem, a fair number of MCMC sampling iterations is needed for a reliable solution. It also requires a separate validation set $A$, which often means that the system is required to split the provided training set into the real training set and the validation set. This makes the transformation set learning itself to use less training data in practice.

Here, we present another approach of using the \emph{entropy} measure. Entropy is an information-theoretic measure that represents the amount of information needed, and it is often used to measure uncertainty (or information gain) in machine learning \cite{settles2010active}. Our idea is to learn the set $S^*$ by iteratively finding transform filters $F_{1 \cdots n}$ that will provide us the maximum amount of information gain when applied to the (original HR) training videos we have.

We formulate the problem similar to the active learning problem. At each iteration, we select $F_k$ that will generate new LR samples with the most uncertainty (i.e., maximum entropy) measured based on the current classifier trained with the current set of transforms: $f_{\theta(S^t)}$. Adding such samples to the training set makes the new classifier to have the most information gain. That is, we iteratively update our set as $S^{t+1} = S^t \cup \{F_*^{t}\}$ where
\begin{equation}
	\begin{aligned}
		F^{t}_* &= \argmax_k \sum_i H(D_kF_kX_i)\\
		    &= \argmax_k - \sum_i \sum_j P_{\theta(S^t)}(y_j | D_kF_kX_i)\\
		    &\relphantom{= \argmax - \sum \sum}\log  P_{\theta(S^t)}(y_j | D_kF_kX_i).
	\end{aligned}
\end{equation}
Here, $X_i$ is each video in the training set, and $P_{\theta(S^t)}$ is the probability computed from the classifier $f_{\theta(S^t)}$. We are essentially searching for the filter that will provide the largest amount of information gain when added to the current transformation set $S^t$. More specifically, we sum the entropy $H$ (i.e., uncertainty) of all low resolution training videos that can be generated with the filter $F_k$: $H(D_kF_kX_i)$.

The approach iteratively adds one transform $F^t_*$ at every iteration $t$, which is the greedy strategy based on the entropy measure, until it reaches the $n$th round: $S^* = S^n$. Notice that such entropy can be measured with any videos with/without ground truth labels. This makes the proposed approach suitable for the unsupervised (transform) learning scenarios as well.

\section{Experiments}

We confirm the effectiveness of inverse super resolution using low resolution version (16x12 and 32x24) of three different datasets (HMDB, DogCentric, and JPL-Interaction).

\subsection{Resized datasets}

We selected three public datasets and resized them to obtain low resolution (e.g., 16x12) videos.

HMDB dataset \cite{kuehne11} is a dataset popularly used for video classification. It is composed of videos with 51 different action classes, containing $\sim$7000 videos. The videos include short clips, mostly from movies, obtained from YouTube. DogCentric dataset \cite{ryoo14dog} and JPL-Interaction dataset \cite{ryoo13} are the first-person video datasets taken with wearable/robot cameras. They are smaller scale datasets, having $\sim$200 videos and $\sim$10 activity classes. DogCentric activity dataset is composed of first-person videos taken from a wearable camera mounted on top of a dog interacting with humans and surroundings. The dataset contains a significant amount of ego-motion, and it serves as a benchmark to test whether an approach is able to capture information in activities while enduring/capturing strong camera ego-motion. JPL-Interaction dataset contains human-robot activity videos taken from a robot's point-of-view. 

\begin{figure}
	\centering
	\resizebox{0.95\linewidth}{!}{
	  \includegraphics{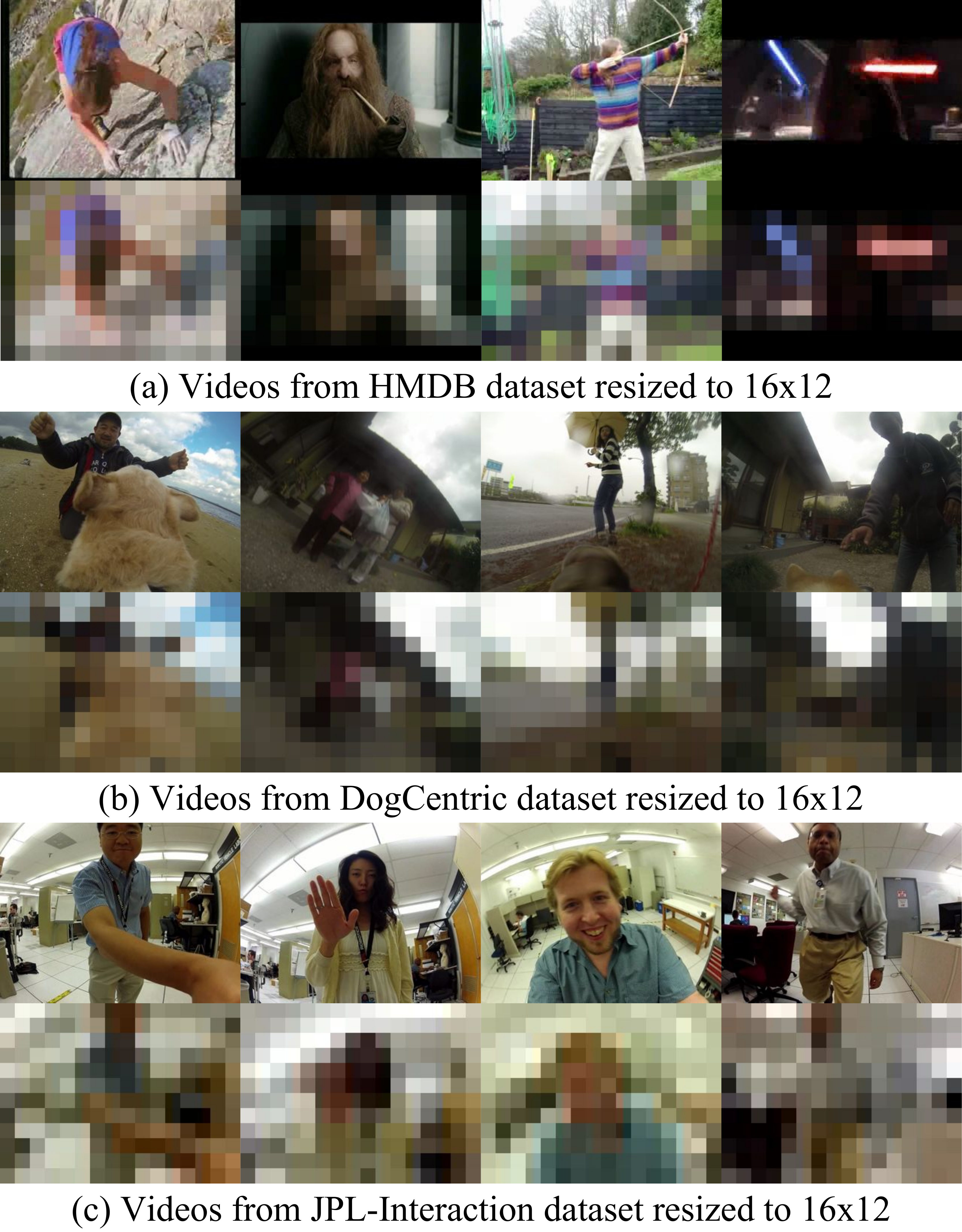}
	}
	\caption{The original resolution videos (top) and their 16x12 resized videos (bottom) from the three datasets used. We can confirm that the videos are properly anonymized (i.e., we cannot distinguish human faces) by resizing them to extreme low resolution, but activity recognition from them is becoming more challenging due to the loss of details.}
	\label{fig:videos}
\end{figure}

We emphasize once more that we made all these videos in the datasets have significantly lower resolution (Figure \ref{fig:videos}), which is a much more challenging setting compared to the original datasets with their full resolution. Our main video resolution setting was 16x12, and we also tested the resolution of 32x24. For the resizing, we used the approach of averaging pixels in the original high-resolution videos that fall within LR pixel boundaries. A video cropping was used for the videos with non-4:3 aspect ratio.

\subsection{Implementation}

\begin{figure}
	\centering
	\resizebox{0.8\linewidth}{!}{
	  \includegraphics{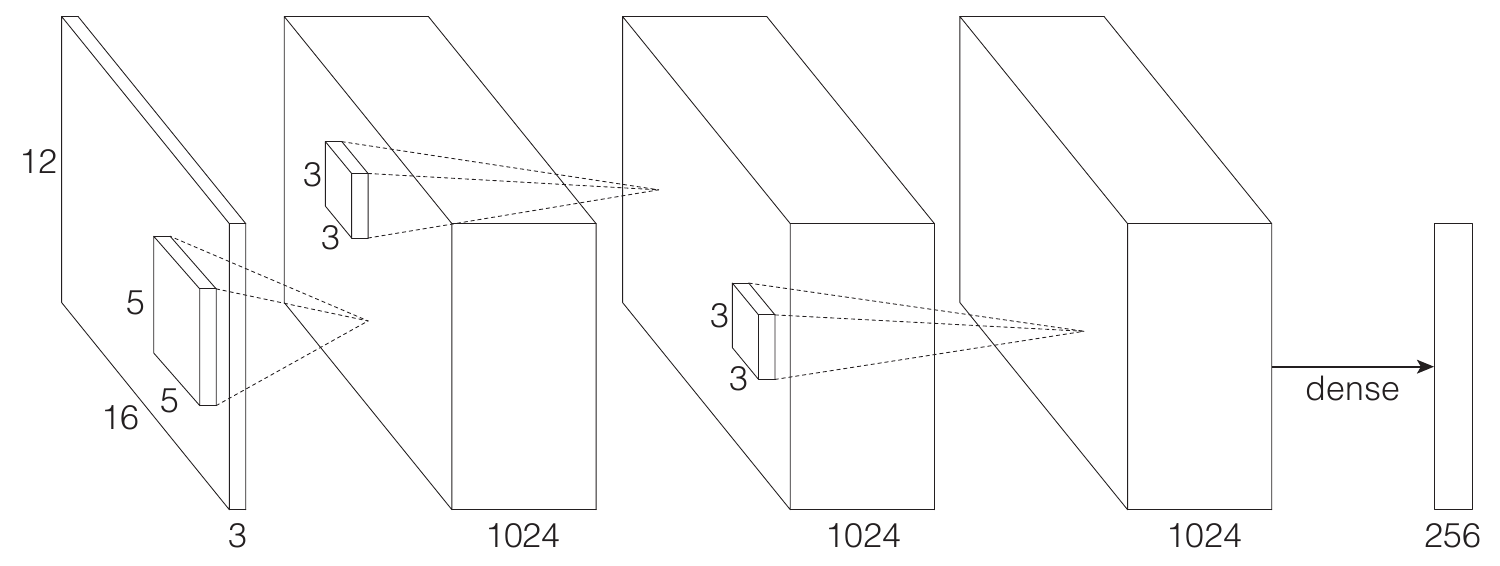}
	}
	\caption{CNN architecture for extracting 256-D features from very low resolution images.}
	\label{fig:LR_cnn}
\end{figure}

{\flushleft\textbf{Feature descriptors/representation:} We extracted three different types of popular video features and tested our inverse super resolution with each of them and their combinations. The three features are (i) histogram of oriented gradients (HOG), (ii) histogram of optical flows (HOF), and (iii) convolutional neural network (CNN) features. These feature descriptors were extracted from each frame of the video. In order to make the CNN handle our low-resolution frames, we newly designed and utilized a 3-layer network with dense convolution and minimal pooling, illustrated in Figure \ref{fig:LR_cnn}. Next, we use Pooled Time Series (PoT) feature representation \cite{ryoo15} with temporal pyramids of level 1 or 4 on top of these four descriptors.}

{\flushleft\textbf{Classifier:} Standard SVM classifiers with three different kernels were used: a linear kernel and two non-linear multi-channel kernels ($\chi^2$ and the histogram-intersection kernels).}


{\flushleft\textbf{Baselines:} The conventional approach for low resolution activity recognition is to simply resize original training videos to fit the target resolution (Figure \ref{fig:isr} (a)). We use this as our baseline, while making it utilize the identical features and representation. The parameters were tuned for each system individually. In addition, we implemented the \emph{data augmentation} (DA) approach similar to \cite{karpathy14} that randomly selects LR transformations to increase the number of training samples. We added random rotation transformations to the data augmentation as well (DA+rotation), and also tested the uniform transformation selection strategy.}

\subsection{Evaluation}

We conducted experiments with the videos downsampled to 16x12 (or 32x24) as described above. We followed the standard evaluation setting for each of the datasets. In the HMDB experiment, we used the provided 3 training/testing splits and performed the 51-class classification. In the experiments with the DogCentric dataset, multiple rounds of random half-half training/test splits were used to measure the accuracy. In the case of JPL-Interaction dataset with robot videos, 12-fold leave-one-set-out cross validation was used.


\begin{figure*}
	\centering
	\resizebox{1.0\linewidth}{!}{
	  \includegraphics{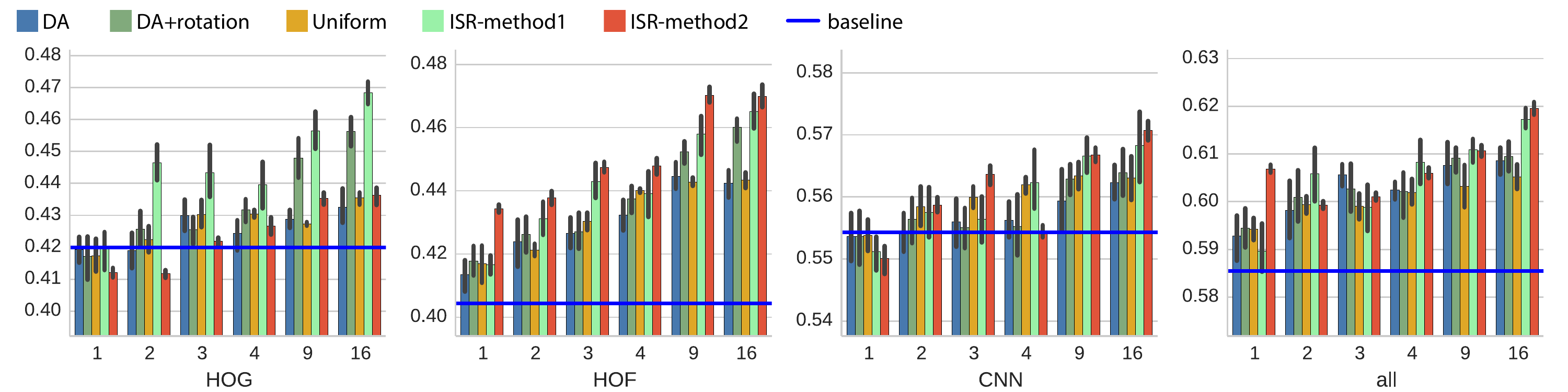}
	}
	\caption{Experimental results with different features on 16x12 DogCentric dataset. X-axis shows the number of LR samples obtained using ISR or data augmentation (i.e., $n$), and Y-axis is the classification accuracy (\%). The blue horizontal line in each plot shows the activity classification performance without ISR. ISR shows a superior performance in all cases.}
	\label{fig:exp-dog}
\end{figure*}

\subsubsection{16x12 DogCentric dataset}
\label{subsubsec:dog}


\begin{table}
    \small
	\caption{Performances (\%) of different methods tested with 16x12 DogCentric dataset, using three different kernels. $n=16$ and PoT level 1 was used with all features. Standard deviations were $\sim0.3$, and the behaviors were very consistent.}
	\label{table:kernels}

	\small
	\center
	\setlength\extrarowheight{0pt}

		\begin{tabular}	{c|c|c|c}
			\hline 	 & ~~Linear~~ & ~~~~$\chi^2$~~~~ & Histogram \tabularnewline
			\hline 	Baseline (PoT) & 58.47	& 63.33	& 58.98  \tabularnewline
			        DA & 60.86	& 63.36	& 62.08  \tabularnewline
			        DA + rotation & 60.94	& 64.17	& 62.85  \tabularnewline
			        Uniform & 60.56	& 63.95	& 62.29  \tabularnewline
			\hline
					ISR-method1  & 61.73	& 64.85	& 63.35  \tabularnewline
					ISR-method2 & \textbf{61.96}	& \textbf{64.91}	& \textbf{63.61} \tabularnewline
			\hline
		\end{tabular}

\end{table}

In this experiment, we used 4 types of features, 3 different types of kernels (i.e., linear, $\chi^2$, and histogram-intersection), and 6 different settings for the number of LR samples generated per original video (i.e., $n$). For each of the settings, five different types of sample generation methods were tested: data augmentation, data augmentation with rotations, uniform sampling, and our ISR transform learning methods (method1 and method2).

\begin{table}
	\caption{16x12 DogCentric dataset result comparison: Notice that \cite{wang13} was not able to extract any trajectories from 16x12 videos. For the PoT and our ISR, we are reporting the result with $\chi^2$ kernel with PoT level 4.}
	\label{table:dogcentric}

	\small
	\center
	\setlength\extrarowheight{0pt}

		\begin{tabular}	{c|c|c}
			\hline 	Approach & Resolution &Accuracy \tabularnewline
			\hline 	Iwashita et al. \shortcite{ryoo14dog} & 320x240	& 	60.5 \%  \tabularnewline
			        Wang and Schmid \shortcite{wang13} & 320x240	& 	67.6 \%  \tabularnewline
			        PoT (Ryoo et al. 2015) & 320x240	& 	73.0 \%  \tabularnewline
			\hline
			        Iwashita et al. \shortcite{ryoo14dog} & 16x12	& 	46.2 \%  \tabularnewline
			        Wang and Schmid \shortcite{wang13} & 16x12	& 	10.0 \%  \tabularnewline
					PoT (HOG + HOF + CNN)   & 16x12 & 	64.6 \%  \tabularnewline
					\textbf{ISR} ($n = 16$)  & 16x12 &  67.4 \% \tabularnewline
			\hline
		\end{tabular}

\end{table}

\begin{table*}
	\caption{HMDB performances with and without our ISR. Classification accuracies (\%) on 16x12 and 32x24 are reported. We show means and standard deviations of repeated experiments. The best performance per feature is indicated with bold.}
	\label{table:exp-hmdb}

	\centering
	\resizebox{1.0\linewidth}{!}{%
	  \includegraphics{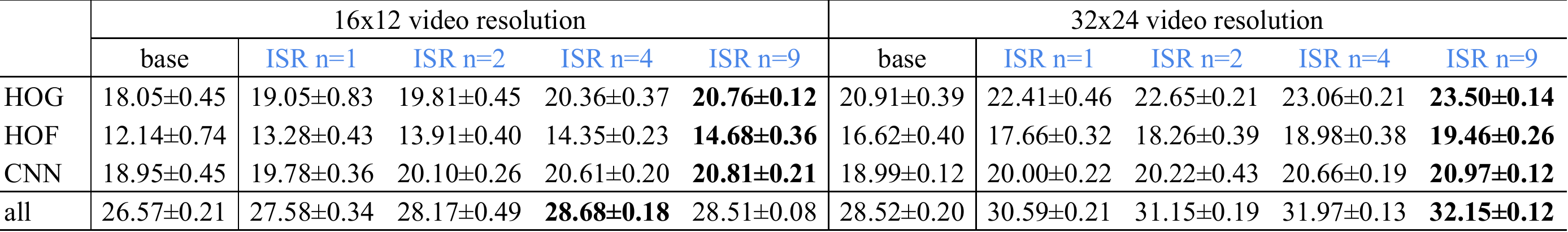}
	}
\end{table*}

Figure \ref{fig:exp-dog} shows the results of our ISR tested with multiple different features while using a linear kernel, and Table \ref{table:kernels} shows the results with three different kernels. We are able to observe that our method1 always performs superior to conventional approaches including data augmentation or uniform transform selection, very reliably. Our method2 performance was less consistent due to that it only uses information gain instead of taking advantage of sample ground truth labels when learning transforms.
Nevertheless, method2 showed an overall performance meaningfully superior to the other approaches (e.g., method2 - 62.0 vs. data augmentation - 60.9 with a linear kernel).

Table \ref{table:dogcentric} compares our approach with other state-of-the-art approaches. The best performance report on the DogCentric dataset is 73\% with 320x240 videos using the PoT feature representation \cite{ryoo15}, but this method only obtains the accuracy of 64.6\% with 16x12 anonymized videos. Our inverse super resolution is further improving it to 67.4\% while using the same features, representation, and classifier.


\subsubsection{Tiny-HMDB dataset: 16x12 and 32x24}




Table \ref{table:exp-hmdb} shows the results with our ISR-method2. The result clearly suggests that inverse super resolution improves low resolution video recognition performances in all cases. Even with a small number of additional ISR samples (e.g., $n = 2$), the performance improved by a meaningful amount compared to the same classifier without inverse super resolution. Naturally, the classification accuracies with 32x24 videos were higher than those with 16x12. CNN performances were similar in both 16x12 and 32x24, since our CNN takes 16x12 resized videos as inputs.

Notice that even with 16x12 downsampled videos, where visual information is lost and the use of trajectory-based features are prohibited, our methods were able to obtain performance superior to several previous methods such as the standard HOF/HOG classifier (20.0\% \cite{kuehne11}) and ActionBank (26.9\% \cite{corso12}). That is, although we are extracting features from 16x12 videos where a person is sometimes as small as a few pixels, it is performing superior to certain methods using the original HR videos (i.e., videos larger than 320x240). The approach \cite{wang13} obtaining state-of-the-art performance of 57.2\% with 320x240 HR videos got the performance of $\sim$2\% in LR videos, since no trajectories could be extracted from 16x12 or 32x24.


\subsubsection{16x12 JPL-Interaction dataset}

We also conducted experiments with the segmented version of JPL-Interaction dataset \cite{ryoo13}, containing robot-centric videos. Figure \ref{fig:videos} (c) shows examples of its 16x12 version, where we can observe that human faces are anonymized.

Table \ref{table:jpl-interaction} shows the results. We are able to confirm once more that the proposed concept of inverse resolution benefits low resolution video recognition. Surprisingly, probably due to the fact that each activity in the dataset shows very distinct appearance/motion different from the others, we were able to obtain the activity classification accuracy comparable to the state-of-the-arts while only using 16x12 extreme low resolution videos.



\section{Conclusion}

We present an \emph{inverse super resolution} method for improving classification performance on extreme low resolution video. We experimentally confirm its effectiveness using three different public datasets. The overall recognition was particularly successful with first-person video datasets, where capturing ego-motion is the most important. Our approach is also computationally efficient in practice, requiring learning iterations linear in the number of ISR samples when using our method 2. In contrast, to achieve similar performance with traditional data augmentation, an order of magnitude more examples are needed (e.g., $n$=$16$ vs. $n$=$175$).

\begin{table}
    \small
	\caption{Recognition performances of our approach tested with 16x12 resized JPL-Interaction dataset. Although our result is based on extremely low resolution 16x12 videos, it obtained comparable performance to the other methods \cite{ryoo13,wang13,narayan14} tested using much higher resolution 320x240 videos.}
	\label{table:jpl-interaction}

	\small
	\center
	\setlength\extrarowheight{0pt}

		\begin{tabular}	{c|c|c}
			\hline 	Approach & Resolution &Accuracy \tabularnewline
			\hline 	Ryoo and Matthies \shortcite{ryoo13} & 320x240	& 	89.6 \%  \tabularnewline
			        Wang and Schmid \shortcite{wang13} & 320x240	& 	96.1 \%  \tabularnewline
			        Narayan et al. \shortcite{narayan14} & 320x240	& 	96.7 \%  \tabularnewline
			\hline
			        Ryoo and Matthies \shortcite{ryoo13} & 16x12	& 	74.5 \%  \tabularnewline
					PoT   & 16x12 & 	92.9 \%  \tabularnewline
					Ours (PoT + \textbf{ISR})  & 16x12 &  96.4 \% \tabularnewline
			\hline
		\end{tabular}

\end{table}

\section{Discussions}

One natural question is whether the resolution in our testing videos is small enough to prevent the human/machine recognition of faces (i.e., whether our videos are really privacy-preserving videos).

\begin{figure}
	\centering
	\resizebox{1.0\linewidth}{!}{
	  \includegraphics{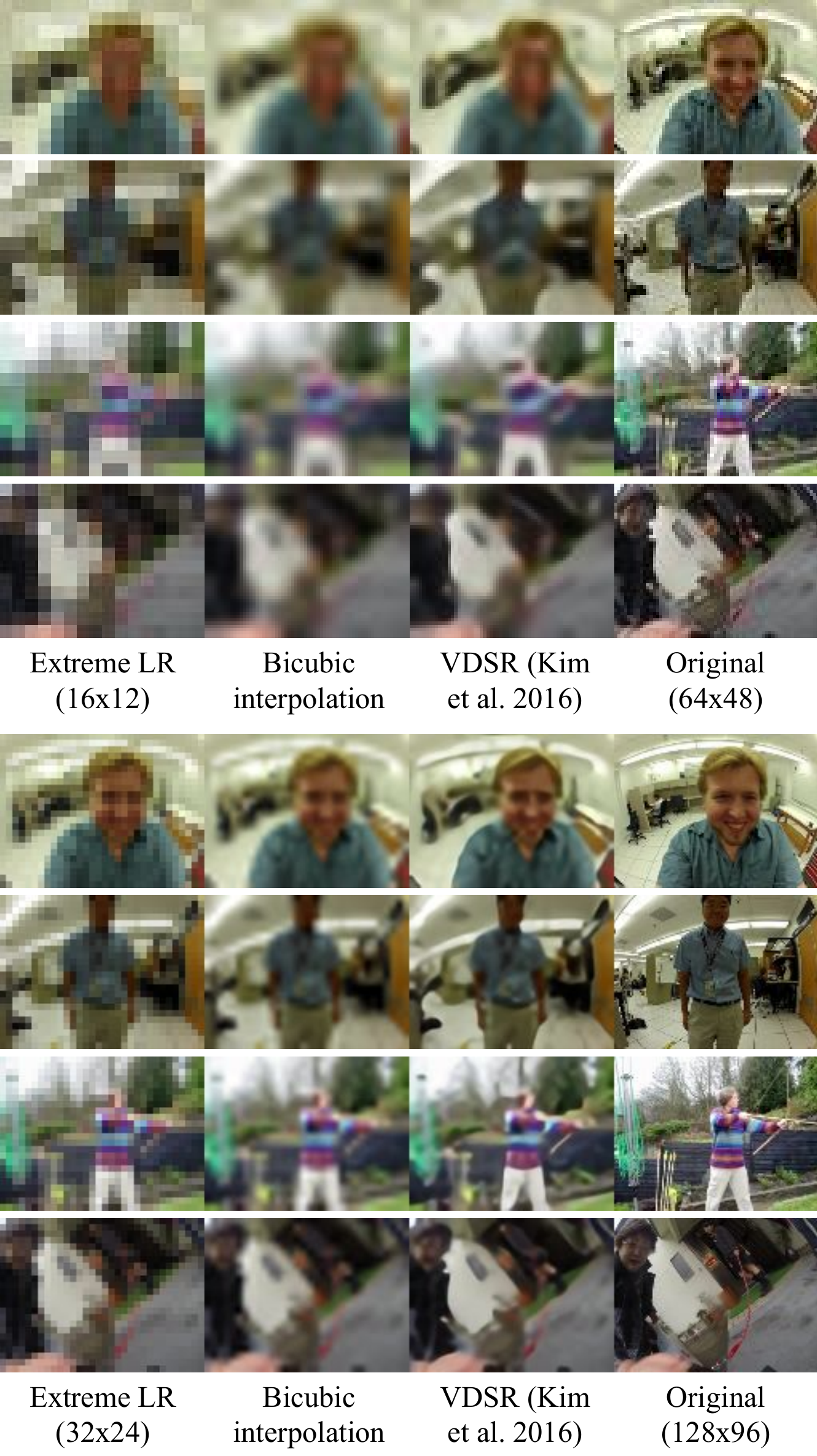}
	}
	\caption{Example resolution enhancement attempts using \cite{kim16sr}. We can observe that face details are not being recovered properly, even after the x4 scale enhancement. This is particularly so with our 16x12 videos.}
	\label{fig:hr-recovery}
\end{figure}

The state-of-the-art low resolution face recognition (i.e., face identification) algorithm using convolutional neural networks \cite{lrface16} obtained around 50$\sim$60\% accuracy with 16x16 human face images. This 50$\sim$60\% classification accuracy is based on the dataset with 180 subjects, and the performance is expected to go even lower in real-world environments with more subjects to be considered for the classification. On the other hand, in our extreme low resolution videos (i.e., 16x12 videos), the resolution of human face is at most 5x7. In often cases, the face resolution was as small as 2x2 or even 1x1. This suggests that reliable face recognition from our extreme low resolution videos will be difficult for both machines and humans. Furthermore, there is a user study \cite{butler15} reporting that anonymizing videos in a way similar to what we have done significantly lowers human's privacy sensitivity.

Another relevant question is whether enhancing the resolution of the testing video (i.e., recovering high resolution faces from LR images) is possible with our extreme LR videos. In order to confirm that such recovery is not possible due to the information loss, we applied the state-of-the-art deep learning-based recovery approach \cite{kim16sr} to the video frames. Figure \ref{fig:hr-recovery} illustrates the results. Notice that these images are based on the scale factor of x4. Any attempt with the higher scale factor gave us worse results. We observe that this deep learning-based resolution enhancement is not recovering the face details, particularly in 16x12 videos. The algorithm sharpens the edges compared to bicubic interpolation, but fails to recover actual details.



\renewcommand{\thefootnote}{\fnsymbol{footnote}}

\section*{Acknowledgement}

Ryoo and Yang's research in this work was conducted as a part of EgoVid Inc.'s research activity on privacy-preserving computer vision. Ryoo and Yang are the corresponding authors.

{\small
\bibliographystyle{aaai}
\bibliography{isr}
}

\end{document}